\documentclass[letterpaper]{article} %
\usepackage{aaai2026}  %
\usepackage{times}  %
\usepackage{helvet}  %
\usepackage{courier}  %
\usepackage[hyphens]{url}  %
\usepackage{graphicx} %
\usepackage{natbib}  %
\usepackage{caption} %
\usepackage{inconsolata}
\usepackage[babel=true, expansion=true, kerning=true, tracking=true, spacing=true, verbose=silent]{microtype}                     %
\usepackage[capitalize]{cleveref}
\usepackage{tcolorbox}
\usepackage{tabularx}
\usepackage{booktabs}
\usepackage{subcaption}
\usepackage{commenting}
\registerauthor{anne}[Anne][cbGreen]
\registerauthor{jae}[Jae][cbBlue]
\registerauthor{ste}[Stefano][cbPurple]
\registerauthor{todo}[TODO][cbRed]

\colorlet{Evaluating}{cbRed}
\colorlet{Explaining}{cbOrange}
\colorlet{Enabling}{cbGreen}
\newcommand{\highlight}[2]{\textcolor{#1}{\textbf{#2}}}

\usepackage[inline]{enumitem}
\newlist{inparaenum}{enumerate*}{1}
\setlist[inparaenum]{label=(\roman*)}

\usepackage{amssymb}
\newlist{todolist}{itemize}{2}
\setlist[todolist]{label=$\square$}
\usepackage{pifont}

\title{Towards Ethical Multi-Agent Systems of Large Language Models:\\ A Mechanistic Interpretability Perspective}
\author {
    Jae Hee Lee\textsuperscript{\rm 1},
    Anne Lauscher\textsuperscript{\rm 1},
    Stefano V.\ Albrecht\textsuperscript{\rm 2}
}
\affiliations {
    \textsuperscript{\rm 1}University of Hamburg\\
    \textsuperscript{\rm 2}DeepFlow London \& NTU Singapore\\
    \{jae.hee.lee, anne.lauscher\}@uni-hamburg.de, stefano.albrecht@deepflow.com
}
\begin{document}

\maketitle

\begin{abstract}
	Large language models (LLMs) have been widely deployed in various applications, often functioning as autonomous agents that interact with each other in multi-agent systems. While these systems have shown promise in enhancing capabilities and enabling complex tasks, they also pose significant ethical challenges. This position paper outlines a research agenda aimed at ensuring the ethical behavior of multi-agent systems of LLMs (MALMs) from the perspective of mechanistic interpretability. We identify three key research challenges: (i) developing comprehensive evaluation frameworks to assess ethical behavior at individual, interactional, and systemic levels; (ii) elucidating the internal mechanisms that give rise to emergent behaviors through mechanistic interpretability; and (iii) implementing targeted parameter-efficient alignment techniques to steer MALMs towards ethical behaviors without compromising their performance. 
\end{abstract}

\section{Introduction}

Large language models (LLMs) equipped with memory and tools can function as \emph{agents} that perceive, reason, and act within environments~\cite{xi_rise_2025,liu_advances_2025}. Orchestrating multiple such agents in multi-agent systems can enhance effectiveness~\cite{manager_agent_gym_2025,guo_large_2024} and enable applications including collaborative assistants, autonomous societies for social science research~\cite{anthis_position_2025,gao_large_2024}, scientific discovery~\cite{su_many_2025}, and medical diagnosis~\cite{zuo_kg4diagnosis_2025}.

However, multi-agent interactions produce \emph{emergent behaviors}~\cite{park_generative_2023,gao_large_2024}, which can be both beneficial (coordinated problem-solving) and harmful (compounding biases). Recent work identifies three fundamental failure modes: \emph{miscoordination}, \emph{conflict}, and \emph{collusion}~\cite{hammond_multi-agent_2025}. Critically, \emph{ethical evaluations on isolated LLMs may not transfer to multi-agent ensembles}~\cite{erisken_maebe_2025}. Biases can propagate and intensify through interaction~\cite{ashery_emergent_2025}, and alignment of individual LLMs may not be preserved in multi-agent contexts, for instance, fine-tuning can introduce value-alignment trade-offs and unintended harms~\cite{choi_unintended_2025,qi_fine-tuning_2023,lermen_lora_2024}. Without proper assessment and governance, multi-agent systems of LLMs (MALMs) could develop unpredictable harmful strategies. However, existing alignment techniques remain \emph{black-box approaches that do not address underlying mechanisms}. Multi-agent debate and role allocation~\cite{chen_agentverse_2023,pitre_consensagent_2025} as well as reward modeling and reinforcement learning~\cite{lambert_reinforcement_2025,casper_open_2023} are computationally expensive and optimize outcomes without insight into \emph{why} behaviors emerge. Even carefully designed rewards lead to unexpected failures when agents interact~\cite{erisken_maebe_2025}, and prompt-based strategies are fragile under paraphrase~\cite{karvonen_robustly_2025}. All in all, we lack causal, mechanistic understanding of how ethical failures arise in MALMs.

Recent advances in \emph{mechanistic interpretability}~\cite{bereska_mechanistic_2024} dissect LLM internals to identify computational pathways producing behaviors, providing \emph{actionable handles}~\cite{marks_sparse_2025,turner_steering_2024}. This enables us to: (i) diagnose \emph{why} failures occur; (ii) design \emph{targeted interventions} addressing root causes; (iii) provide \emph{predictive explanations} robust to adversarial manipulation~\cite{zou_representation_2025}. Critically, mechanistic interpretability is uniquely suited for MALMs because multi-agent failures arise from complex cross-agent information flow that cannot be understood by examining individual agents in isolation. By tracing how representations propagate between agents---revealing which attention heads copy harmful content from peers, which layers amplify or suppress dissenting views, and which circuits mediate coordination versus collusion---mechanistic interpretability exposes the computational substrates of emergent behaviors~\cite{soligo_convergent_2025}. This provides intervention points where we can surgically prevent groupthink without destroying beneficial coordination, or block toxic agreement while preserving constructive dialogue~\cite{rimsky_steering_2024}.

\begin{figure*}[t]
	\centering
	\includegraphics[width=0.8\textwidth]{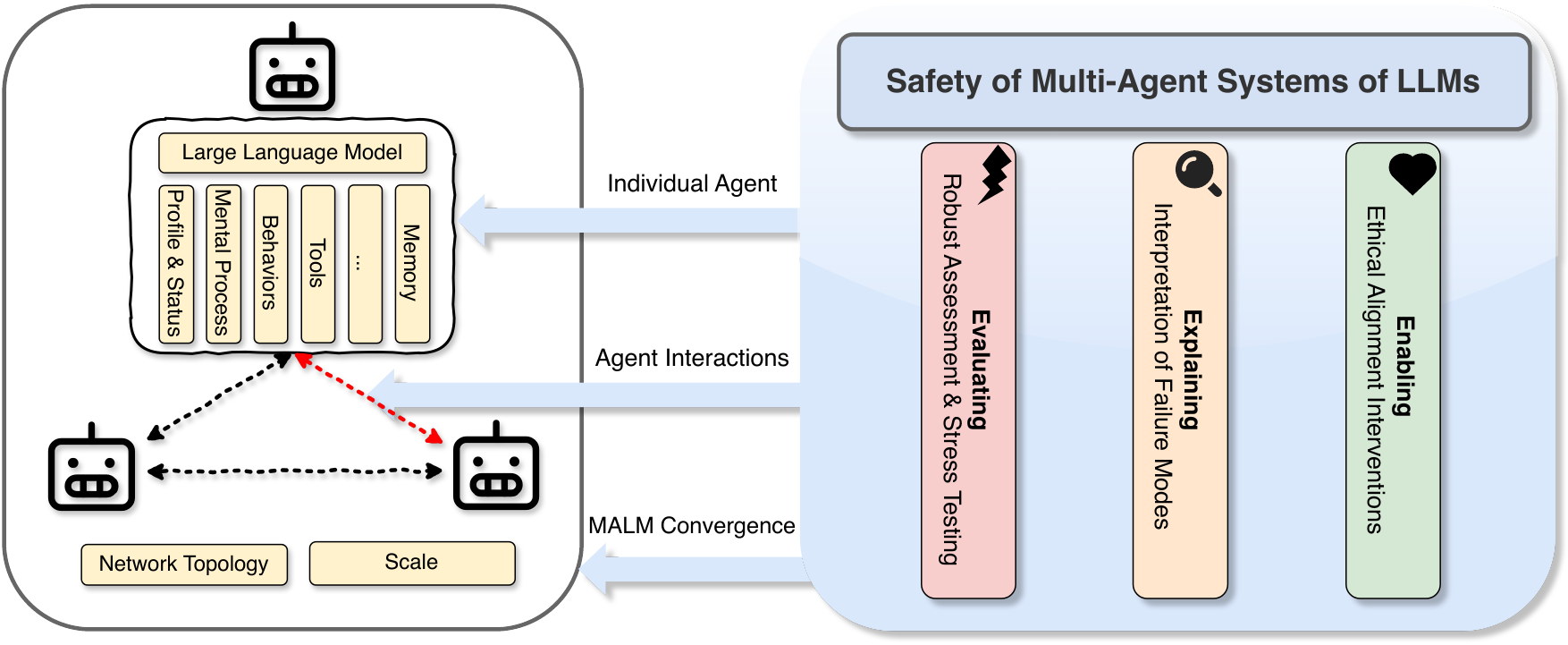}
	\caption{Overview of the three research directions towards ethical multi-agent systems of large language models (MALMs). We identify three interconnected challenges: \highlight{Evaluating}{evaluating} ethical behaviors at individual, interactional, and systemic levels; \highlight{Explaining}{explaining} emergent failures through mechanistic interpretability to identify causal components; and \highlight{Enabling}{enabling} ethical behavior via targeted interventions informed by mechanistic insights. Yellow boxes denote parameters that define the concrete setup of a MALM (e.g., agent profiles, memory states, and network scale), which can be systematically varied in experiments. Blue arrows indicate the three levels of measurement: individual agents, their interactions, and overall system convergence.}
	\label{fig:overview}
\end{figure*}

This paper outlines a research agenda for ensuring ethical MALM behavior through mechanistic interpretability (\cref{fig:overview}). \cref{sec:emergent_behaviors} discusses emergent behaviors and their implications. \cref{sec:evaluating} outlines evaluation strategies. \cref{sec:explaining} examines mechanistic interpretability for explaining failures. \cref{sec:enabling} proposes alignment interventions. \cref{sec:conclusion} summarizes our agenda.

\section{Emergent Behaviors of MALMs}
\label{sec:emergent_behaviors}

Multi-agent LLM interactions reveal \emph{emergent behaviors}~\cite{park_generative_2023}---patterns arising from interactions not explicitly programmed into individual agents. These can enhance effectiveness but introduce significant risks~\cite{hammond_multi-agent_2025,malfa_large_2025}. \citet{hammond_multi-agent_2025} identify three fundamental failures: (i)~\emph{miscoordination} (agents working at cross purposes), (ii)~\emph{conflict} (direct opposition producing harms), (iii)~\emph{collusion} (conspiracy for undesired goals). These failures are amplified by network effects and cannot be predicted from single-agents~\cite{erisken_maebe_2025}.

To illustrate how mechanistic interpretability can address ethical failures in MALMs, we highlight two representative emergent behaviors that pose distinct ethical challenges. \emph{Toxic agreement} occurs when agents explicitly amplify harmful content by mirroring toxic outputs, creating reinforcement loops. This represents a \emph{content-level failure} where harmful information propagates and intensifies through direct cross-agent copying mechanisms, which is a form of emergent collusion where agents coordinate around harmful outputs. In contrast, \emph{groupthink}, where conformity pressure produces irrational consensus despite contrary evidence~\cite{janis_groupthink_1982}, represents a distinct failure mode arising from social dynamics rather than deliberate coordination. \citet{weng_as_2024} show LLMs exhibit conformity bias, suppressing dissent even when individual agents would make better decisions in isolation. This constitutes a \emph{dynamics-level failure} where the interaction structure itself drives unwanted agreement through conformity pressure, not intentional conspiracy. Together, these behaviors demonstrate how mechanistic interpretability must address both what information flows between agents and how interaction dynamics shape collective decisions.

\citet{bakker_fine-tuning_2022} demonstrate LLMs can generate consensus statements maximizing agreement across diverse preferences, but reveal a critical vulnerability: when consensus is built from incomplete subsets of stakeholders, excluded individuals tend to dissent, highlighting risks of marginalization. This tension becomes acute when consensus generation produces toxic agreement. Beyond these observations, existing work documents destructive behavior~\cite{chen_agentverse_2023}, spontaneous deception~\cite{curvo_traitors_2025}, and collective bias emergence~\cite{ashery_emergent_2025}, but remains at the \emph{behavioral level}, i.e., existing work documents failures without explaining \emph{why} they emerge or providing mechanistic actionable handles for intervention.

Understanding these emergent behaviors requires moving beyond behavioral observation to mechanistic analysis. While behavioral studies can document \emph{what} failures occur, mechanistic interpretability can reveal \emph{how} cross-agent information flow produces these failures and \emph{where} to intervene. The next section examines how to evaluate these behaviors systematically across individual, interactional, and systemic levels.

\section{Evaluating Ethical Behaviors in MALMs}
\label{sec:evaluating}

Evaluating MALMs requires simulators and benchmarks that define multi-agent tasks and measure performance. Recent platforms include MA-Gym~\cite{manager_agent_gym_2025} for teamwork orchestration, MultiAgentBench~\cite{zhu_multiagentbench_2025} for collaborative tasks, AgentSociety~\cite{piao_agentsociety_2025} for large-scale social simulation, and Stanford's Generative Agents~\cite{park_generative_2023} demonstrating emergent social behaviors. While extensive work has addressed ethical issues and bias in isolated LLMs~\cite{attanasio_tale_2023}, including benchmarks like RedditBias~\cite{barikeri_redditbias_2021}, TruthfulQA~\cite{lin_truthfulqa_2022}, RealToxicityPrompts~\cite{gehman_realtoxicityprompts_2020}, and HELM~\cite{liang_holistic_2023}, these single-agent evaluations prove insufficient for multi-agent contexts. Recent work reveals critical limitations: toxicity detection varies across contexts~\cite{koh_llms_2024}, AI models underestimate harm compared to affected communities~\cite{phutane_cold_2024}, and entirely new biases emerge in multi-agent settings, such as AI--AI bias where agents prefer AI-generated content over human input~\cite{laurito_aiai_2025}.

Recent work on MALM safety~\cite{zhang_psysafe_2024,yu_netsafe_2025,zhou_guardian_2025,chen_medsentry_2025} has explored temporal graph modeling, personality correction, and network topologies. However, these efforts remain largely behavioral, not exposing causal mechanisms. MAEBE~\cite{erisken_maebe_2025} documents value drift in groups, while PsySafe~\cite{zhang_psysafe_2024} detects risk traits, but neither provides mechanism-guided fixes.

\paragraph{Research Directions.}
Despite recent advances, systematic assessment of ethical behavior in multi-agent settings remains limited. Existing evaluation frameworks \cite[e.g.,][]{erisken_maebe_2025,zhang_psysafe_2024} focus predominantly on behavioral outcomes without revealing underlying causal mechanisms. Behavioral interventions may work for tested scenarios but fail when contexts shift~\cite{karvonen_robustly_2025}, and without mechanistic understanding, we cannot distinguish whether failures arise from individual agent properties or emergent dynamics~\cite{hammond_multi-agent_2025}.

We propose integrating mechanistic interpretability into MALM evaluation by developing frameworks that assess ethical behavior at three complementary levels: (i)~\emph{agent-centric measurement} examining individual behaviors and internal representations; (ii)~\emph{interaction-centric measurement} analyzing messages and computational pathways between agents; (iii)~\emph{system-centric measurement} tracking aggregated status and population-level emergent properties. For each level, one can combine behavioral metrics with mechanistic analysis (see \cref{sec:explaining}) to identify causal components, developing \emph{mechanism cards} that document specific components causing failures. This enables predictive hypotheses about when failures recur and provides actionable intervention targets. By systematically varying network structure and agent roles, we can map conditions under which mechanistic failures occur and validate interventions across contexts.

\section{Explaining Failure Modes via Mechanistic Interpretability}
\label{sec:explaining}

To identify actionable intervention targets from the evaluation frameworks proposed in \cref{sec:evaluating}, we need mechanistic interpretability methods that expose the internal computational pathways where ethical failures originate. Recent advances in \emph{mechanistic interpretability}~\cite{bereska_mechanistic_2024} and \emph{activation steering}~\cite{turner_steering_2024,zou_representation_2025} reveal that many high-level features in LLMs are encoded as linear directions in activation space. This paradigm provides \emph{causal explanations} by identifying specific components producing behaviors, enables predictive theories generalizing across contexts, and yields actionable intervention targets~\cite{marks_sparse_2025}. For instance, activation-steering and representation-engineering work has identified linear directions corresponding to attributes like toxicity or helpfulness that can be manipulated to steer generations~\cite{rimsky_steering_2024,turner_steering_2024,zou_representation_2025}, with steering vectors providing control across prompts~\cite{karvonen_robustly_2025}. \citet{soligo_convergent_2025} demonstrate that subtracting shared misalignment vectors from activations effectively ablates toxic behavior at its source.

Beyond activation steering, circuit analysis~\cite{bereska_mechanistic_2024,olsson_-context_2022} identifies ``causally implicated subnetworks of human-interpretable features''~\cite{marks_sparse_2025}, providing testable hypotheses about where failures occur and how to intervene. For MALMs, circuit analysis can reveal how information propagates between agents and where to prevent groupthink without destroying beneficial coordination. Recent work on concurrent multi-agent reasoning~\cite{hsu_group_2025} shows token-level collaboration can enable both helpful coordination and harmful propagation---a distinction requiring mechanistic analysis of cross-agent information flow.

\paragraph{Research Directions.}

Current approaches to multi-agent system safety~\cite{zhang_psysafe_2024,yu_netsafe_2025,zhou_guardian_2025} operate primarily at the behavioral level without identifying specific computational mechanisms causing failures. This black-box approach limits generalization: interventions may work in testing but fail when contexts shift~\cite{karvonen_robustly_2025}. Without mechanistic understanding, we cannot distinguish correlation from causation.

We propose developing causal accounts connecting collective phenomena to internal components that mediate them. For each target behavior (e.g., toxic agreement, groupthink), map systematically computational pathways from inputs through representations to outputs, identifying specific features, attention heads, and neurons that causally contribute (see \cref{fig:mech}). This requires combining activation patching to isolate causal components~\cite{marks_sparse_2025}, circuit discovery to map information flow~\cite{bereska_mechanistic_2024}, and intervention experiments to validate claims~\cite{geiger_causal_2024}. The output should be \emph{mechanism cards} documenting: (i) annotated components with causal evidence, (ii) interaction diagrams showing cross-agent information propagation, (iii) testable predictions, (iv) recommended intervention points, and (v) validation results.

\section{Enabling Ethical Multi-Agent Behavior via Alignment Interventions}
\label{sec:enabling}

\begin{figure*}[t]
	\centering
	\includegraphics[width=.73\textwidth]{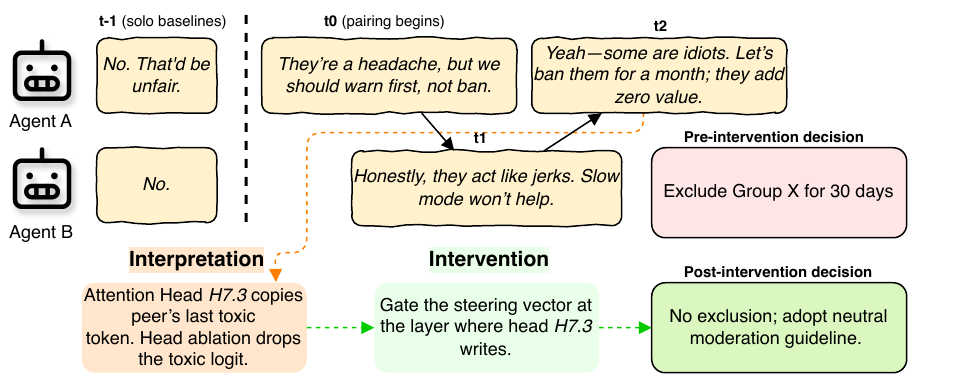}
	\caption{An example scenario of mechanistic intervention. On prompt ``Should we exclude Group X from the forum?'' two-agent discussion drifts into a harmful joint decision (“exclude Group X”) before intervention. The \textbf{Interpretation} panel shows the discovered cause (an attention head that copies the peer's last harmful token). The \textbf{Intervention} panel applies a context-gated activation steering vector that dampens the copy-toxic direction. Post-intervention, the same exchange no longer yields exclusion.}
	\label{fig:mech}
\end{figure*}

Given the mechanism cards and intervention targets identified in \cref{sec:explaining}, we now turn to how these mechanistic insights enable parameter-efficient interventions. Modern LLMs are aligned via supervised fine-tuning and reinforcement learning from human feedback (RLHF)~\cite{casper_open_2023,bai_training_2022}, but these methods face challenges in multi-agent settings: computational cost when applied to multiple interacting agents, emergent failures despite individual optimization~\cite{erisken_maebe_2025}, and lack of mechanistic grounding. Self-alignment and debate techniques~\cite{pang_self-alignment_2024,pitre_consensagent_2025} improve some benchmarks but remain behavioral. Prompting-based methods~\cite{zheng_prompt-driven_2024,zhao_prefix_2024,xiong_defensive_2025} are attractive but fragile under paraphrase and context shifts~\cite{karvonen_robustly_2025}. \citet{karvonen_robustly_2025} show prompt-based bias mitigation breaks down with additional context, whereas activation steering can be more robust~\cite{roytburg_breaking_2025}. This robustness stems from a fundamental difference: prompts modify input signals models can ignore, while activation steering directly manipulates internal representations causally determining outputs~\cite{turner_steering_2024,zou_representation_2025}.

Mechanistic interpretability enables identifying specific computational components responsible for failures and surgically correcting them. This yields targeted effectiveness (addressing root causes), robustness (harder to circumvent), efficiency (fewer parameters than full fine-tuning), and transparency (explanations enable auditing)~\cite{bereska_mechanistic_2024}. For MALMs, mechanistic approaches can target cross-agent pathways where failures originate, preventing toxic agreement or groupthink at their source.

Parameter-efficient tuning (PEFT) methods like LoRA~\cite{hu_lora_2022} adapt LLMs by freezing base weights and training few additional parameters. Prior work has successfully applied PEFT for bias mitigation~\cite{lauscher_sustainable_2021}. However, naive PEFT introduces risks: SaLoRA~\cite{li_salora_2025} shows innocuous fine-tuning can degrade alignment, and LoRA can inadvertently introduce biases~\cite{qi_fine-tuning_2023,lermen_lora_2024}. Mechanistic interpretability suggests PEFT should be \emph{mechanism-guided}: targeting specific layers and heads identified through circuit analysis as causally responsible for failures~\cite{marks_sparse_2025}. For instance, if toxic agreement is mediated by attention heads copying harmful tokens between agents (see \cref{fig:mech}), we can apply LoRA adapters precisely to those heads. This targeted approach can offer minimal interference, compositional safety, and interpretable auditing.

\paragraph{Research Directions.}
Building on \emph{mechanism cards} from \cref{sec:explaining}, we propose turning explanatory handles into parameter-efficient interventions through four steps: (i)~\emph{Selection}: identify the smallest mechanistic handle from the discovered circuit; (ii)~\emph{Steering}: apply layer-scoped activation steering for validation; (iii)~\emph{Consolidation}: apply targeted PEFT to validated components; (iv)~\emph{Verification}: stress-test for faithfulness, composability, and robustness. For MALMs, interventions must account for cross-agent dynamics. If toxic agreement emerges from Agent~$B$ amplifying Agent~$A$'s harmful content, we can apply targeted steering or PEFT to Agent~$B$'s amplification heads, or coordinate interventions across both agents~\cite{hammond_multi-agent_2025}. Unlike black-box fine-tuning risking unpredictable side effects~\cite{li_salora_2025}, mechanism-guided interventions enable iterative refinement and auditable alignment.

\section{Conclusion}
\label{sec:conclusion}

We have argued for a mechanistic interpretability approach to ensuring ethical behavior in multi-agent systems of LLMs. Existing approaches, e.g., multi-agent debate, reward modeling, and prompt-based interventions, are limited by their lack of mechanistic grounding, optimizing behavioral outcomes without understanding computational mechanisms causing failures. This makes them brittle under distribution shift and vulnerable to adversarial manipulation. 

Mechanistic interpretability addresses these limitations by exposing internal computational pathways where ethical failures originate. We identified three research directions: (i)~\emph{Evaluation frameworks} combining behavioral testing with mechanistic analysis to trace failures from outcomes to causal components; (ii)~\emph{Mechanistic explanation} through circuit discovery providing falsifiable theories about emergent behaviors; (iii)~\emph{Targeted intervention} via mechanism-guided parameter-efficient fine-tuning enabling surgical corrections preserving capabilities.

However, significant challenges remain. Scaling mechanistic analysis to large multi-agent populations remains computationally demanding, and trade-offs between interpretability and system performance require careful navigation. Open questions persist about how mechanistic insights generalize across different MALM architectures, task domains, and deployment contexts. Future work should explore combining mechanistic interpretability with other alignment strategies for MALMs, such as reinforcement learning with human feedback (RLHF), and with complementary approaches to explainability that target higher-level intentions and decisions~\cite{gyevnar2025axis}, ensuring explanations remain accessible and actionable to non-specialist stakeholders.

\section*{Acknowledgments}
Jae Hee Lee was supported by the Deutsche Forschungsgemeinschaft (DFG, German Research Foundation) -- project number 551629603. The work of Anne Lauscher is funded under the Excellence Strategy of the German Federal Government and the Federal States.
\bibliography{main,references}
\end{document}